\documentclass[9pt, conference]{IEEEtran}
\IEEEoverridecommandlockouts

\usepackage{cite}
\usepackage{amsmath,amssymb,bm,amsfonts}
\usepackage{algorithmic}
\usepackage{graphicx}
\usepackage{textcomp}
\usepackage{xcolor}
\usepackage{multirow}
\usepackage{threeparttable}
\RequirePackage{graphicx}
\RequirePackage{textcomp}
\RequirePackage{booktabs}
\RequirePackage{siunitx}
\RequirePackage{xspace}
\RequirePackage{url}
\RequirePackage{lipsum}  
\RequirePackage{tipa}
\RequirePackage{enumitem}
\def\BibTeX{{\rm B\kern-.05em{\sc i\kern-.025em b}\kern-.08em
    T\kern-.1667em\lower.7ex\hbox{E}\kern-.125emX}}
\begin{document}

\title{Leveraging Chain of Thought towards Empathetic Spoken Dialogue without Corresponding Question-Answering Data
\thanks{This work is supported by National Key Research and Development Program of China (2021ZD0112900), National Natural Science Foundation of China (62076144) and the Major Key Project of PCL (PCL2023A09).}
}

\author{
\emph{Jingran Xie$^{1,2}$\thanks{* Corresponding author.}, Shun Lei$^{1}$, Yue Yu$^{2}$, Yang Xiang$^{2}$, Hui Wang$^{2}$, Xixin Wu$^{3}$, Zhiyong Wu$^{1,3,*}$}\\
  
    $^1$Shenzhen International Graduate School, Tsinghua University, Shenzhen, China \\
    $^2$Pengcheng Laboratory, Shenzhen, China \\
    $^3$The Chinese University of Hong Kong, Hong Kong SAR, China \\
}

\maketitle
\begin{abstract}
Empathetic dialogue is crucial for natural human-computer interaction, allowing the dialogue system to respond in a more personalized and emotionally aware manner, improving user satisfaction and engagement. The emergence of large language models (LLMs) has revolutionized dialogue generation by harnessing their powerful capabilities and shown its potential in multimodal domains. Many studies have integrated speech with text-based LLMs to take speech question as input and output text response. However, the lack of spoken question-answering datasets that include speech style information to supervised fine-tuning (SFT) limits the performance of these systems. As a result, while these systems excel at understanding speech content, they often struggle to generate empathetic responses. In response, we propose a novel approach that circumvents the need for question-answering data, called Listen, Perceive, and Express (LPE). Our method employs a two-stage training process, initially guiding the LLM to listen the content and perceive the emotional aspects of speech. Subsequently, we utilize Chain-of-Thought (CoT) prompting to unlock the model's potential for expressing empathetic responses based on listened spoken content and perceived emotional cues. We employ experiments to prove the effectiveness of proposed method. To our knowledge, this is the first attempt to leverage CoT for speech-based dialogue.

\end{abstract}

\begin{IEEEkeywords}
Spoken dialog, Large language model, Chain-of-thought.
\end{IEEEkeywords}

\vspace{-1mm}
\section{Introduction}
\label{introduction}
Empathetic dialogue aims to generate personalized responses that not only address the content of the conversation but also recognize the state of users, such as their emotions. 
In this process, speech, as a carrier of both content and emotion, is of paramount importance. With the recent developments in Large Language Models (LLMs) such as GPT \cite{radford2019language, brown2020language} and LLaMA \cite{touvron2023llama1, touvron2023llama2}, these models have demonstrated significant potential as dialogue systems and have been widely applied in multimodal domains \cite{huang2023language, liu2024visual, liu2023llava}, offering new opportunities for empathetic spoken dialogue.

Since text-based LLMs still struggle with controlling speech synthesis, research \cite{zhang2023speechgpt, shu2023llasm, chu2023qwen, tang2023salmonn, wang2023blsp, wang2024blsp} often focuses on speech-to-text (S2T) dialogue systems first.
A common approach is the cascaded pipeline, which combines automatic speech recognition (ASR), speech emotion recognition (SER), and LLMs to generate responses. While this method yields high-quality responses, it suffers from issues like error propagation between modules and high latency (average 5.4s) \cite{openai2024gpt4o}, degrading user experience. To address this, researchers have explored end-to-end (E2E) models that integrate speech and text through supervised fine-tuning (SFT) for smoother human-computer interaction. However, these E2E models struggle to generate emotion-based responses due to the lack of spoken empathetic question-answer (QA) datasets.
Although using TTS to generate dialogue data is a possible solution \cite{xue2023chat, lin2024advancing}, the gap between synthesized and natural speech data still limits its real-world applicability.

To address the lack of natural speech QA dataset, we draw inspiration from the success of Chain-of-Thought (CoT) reasoning \cite{wei2022chain} in both text-based dialogue \cite{chen2023soulchat, chen2024cause} and speech tasks \cite{du2024cot, gong2024seamlessexpressivelm}. CoT prompting breaks complex tasks into smaller, manageable sub-tasks, allowing the model to solve them step-by-step. Therefore, we propose a novel method called ``Listen, Perceive, and Express" (LPE) that aims to eliminate the need for empathetic speech QA data with CoT prompt.
Our approach consists of a two-stage training process that trains LLM to \textbf{listen} the content and \textbf{perceive} emotion of speech separately. Then, we employ Chain of Thought (CoT) \cite{wei2022chain} prompting to enable step-by-step reasoning for \textbf{expressing} empathetic responses based on speech content and emotional cues. Following this LPE framework, our model acquires the capability to generate empathetic responses without requiring SFT with QA data. Although our method still generates text-based responses rather than spoken ones, it provides a low-cost solution for advancing empathetic spoken dialogue. Our contribution can be summarized as following:

\begin{itemize}
    \item We propose LPE framework that introduces CoT into empathetic spoken dialogue. By decomposing empathetic dialogue into listening, perceiving, and expressing steps, we leverage CoT to guide the model step-by-step from known tasks to unseen targets.
    \item We develop a two-stage approach to adapt LLMs for empathetic dialogue, utilizing existing ASR and SER data to replace the need for spoken empathetic QA data and ensure the model learns from natural speech data distribution.
    \item We conduct quantitative evaluations on our LPE and explore the impact of CoT prompt on model performance. The experimental results demonstrate the overall effectiveness of our approach and confirm that CoT significantly enhances the model's ability to generate accurate and empathetic responses.
\end{itemize}

\begin{figure*}[h]
  \centering
  \includegraphics[width=\linewidth]{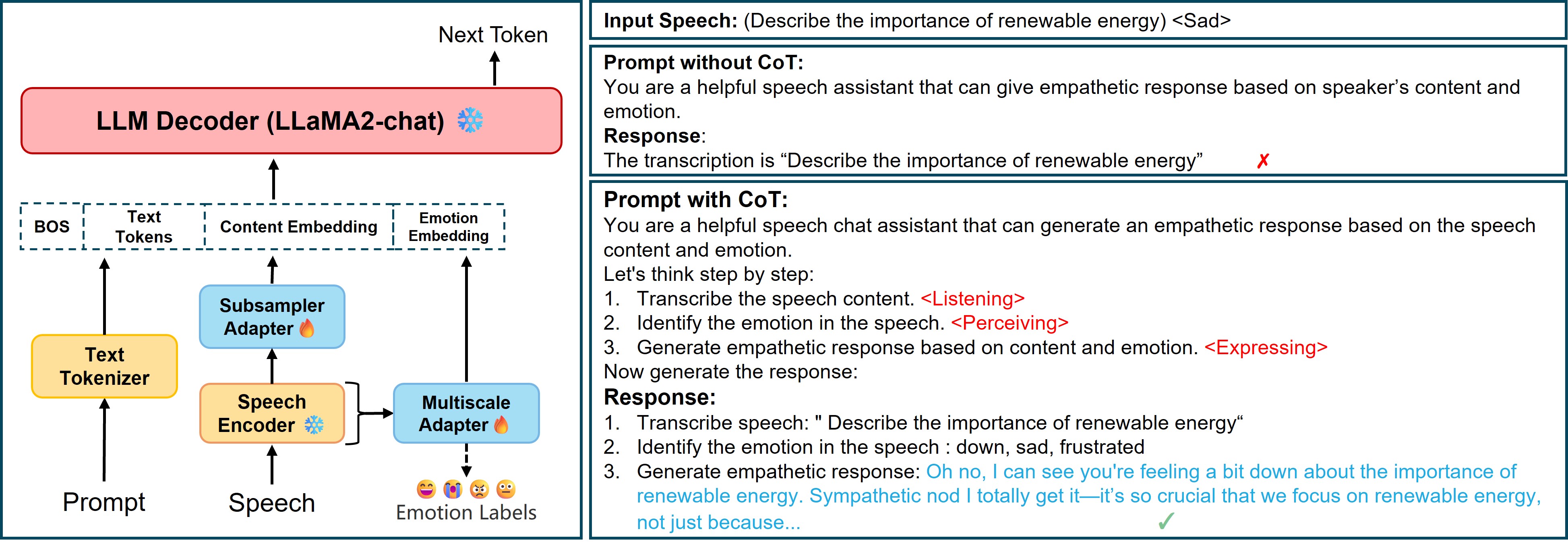}
  \caption{Architecture of the proposed model is shown on left side. On the right is the real sample showing how it listens, perceives, and expresses. () denotes the transcription of the input speech. $<>$ denotes the emotion of the speech. We use \textcolor{red}{$<>$} to highlight each step of LPE, which is not a part of our prompt.}
  \label{fig:model}
  \vspace{-3mm}
\end{figure*}

\section{Proposed Method}
\label{method}
The LPE framework is illustrated in Fig.~\ref{fig:model}. Firstly, we leverage a two-stage training process to train LLM to listen content and perceive emotion of speech. Then, we apply CoT prompting to allow the model to further express empathetic responses based on listening and perceiving without the need for SFT on corresponding QA datasets.

\subsection{Model Architecture}
\label{model_arch}
Our LPE framework consists of three core components: a speech encoder, two feature adapters, and a LLM decoder. We leverage LLaMA’s tokenizer to convert text prompt into text tokens.

\textbf{Speech Encoder} 
We use the WavLM \cite{chen2022wavlm} large model as the speech encoder, which is capable of extracting both content and emotional features from speech. The model consists of 24 Transformer layers, where the upper layers primarily capture content information, and the lower layers focus on paralinguistic features \cite{chen2022wavlm}. We obtain content embedding from the last layer and emotion embedding by applying a weighted sum across output of all layers.

\textbf{Subsampler Adapter} 
Since speech embedding are typically much longer than text embedding, we introduce a subsampler adapter to reduce the length discrepancy between the two modalities. The adapter uses three 1D convolutional layers followed by a bottleneck layer and layer normalization to map the speech features into the textual space.

\textbf{Multiscale Adapter} 
To effectively capture emotional features, multiscale adapter applies weighted sum over all layer outputs of speech encoder. This is followed by two linear layers and mean pooling to generate emotion embedding. A linear classifier is used during training to predict emotion labels, though is not utilized during inference.

\textbf{LLM Decoder} 
To generate text responses, we employ LLaMA2-7B-chat \cite{touvron2023llama2} as the LLM decoder. This model is chosen for its strong conversational abilities. To prevent catastrophic forgetting on internal knowledge, its parameters are frozen throughout the training process.

\subsection{Two-Stage Training}
LPE employs a two-stage training process to achieve the capability of listening speech content and perceiving emotions.

\textbf{Semantic Alignment} 
The goal of the first stage is to align the speech content embedding with the text embedding required by the LLM decoder. We use ASR data to train the subsampler adapter, mapping the speech features into the textual space. The ASR task prompt is passed to the LLaMA's tokenizer to get the text tokens. The input to the LLM decoder consists of a text prompt, the content embedding, and a placeholder emotion embedding. We use a random embedding that serves as an emotion embedding to avoid the model learning information from it in stage 1. During this stage, the model is trained to predict the transcriptions of the input speech using Cross-Entropy (CE) loss.

\textbf{Emotion Alignment} 
Once the subsampler adapter can map speech features into the text space, the LLM is trained to perceive emotional cues. In this stage, we use SER dataset to align emotion labels with LLM embedding space.
Multitask learning \cite{radford2019language, chen2024multi} can improve the training efficiency by reducing overfitting and improve the performance on low-resource tasks, which is popular in NLP and computer vision areas.
We employ multitask learning by assigning each batch a task from ASR, SER, or a combination of ASR and SER at a predefined rate. We use GPT-4o to generate several response templates that contain placeholders for transcriptions and emotion labels which can be filled based on the specific task. Therefore, these three tasks are unified into the text generation task.
Additionally, we adopt continual learning strategy to replay 20\% of the ASR data from stage 1 to avoid  catastrophic forget of knowledge acquired during the first stage. 
In stage 2, we train subsampler and multiscale adapter together to improve the compatibility between the content embedding and emotion embedding, enhancing the ability of the LLM to effectively process both. The CE loss is used to predict response templates, allowing LLM to comprehend both the content and emotional aspects of speech.
Moreover, we apply an additional CE-based emotion classification loss to enhance emotion alignment, ensuring the model to better capture emotional nuances. To be mentioned, the classification result is only used for training, and will not be fed to LLM decoder. The loss in stage 2 is defined as:
\begin{equation}
    Loss = L_{decoder} + \lambda \cdot L_{emotion}
\end{equation}
where $\lambda$ is the hyper parameter and set to 0.1 based on preliminary experiments.

\subsection{Chain of Thought}
\label{cot}
Through the two stage training, LLM is able to understand the content and emotional information in speech. However, as the model has not encountered QA data during training, it struggles to generate proper response, often outputting transcriptions or emotion labels instead. 
To overcome this issue, we introduce the Chain-of-Thought (CoT) \cite{wei2022chain, zhang2022automatic} prompting methodology. CoT improves the model’s reasoning capabilities by breaking down complex tasks into smaller sub-problems. We leverage this capability by splitting the complex task into trainable sub-tasks, such as ASR and SER. We manually design CoT prompt to guide the model follows the step that first transcribes speech (listening), then identifies the emotion of speech (perceiving), and finally generates a response (expressing) that is both contextually and emotionally appropriate base on the transcription and identified emotion. We compare between three types of prompts: 

\textbf{No CoT}
Prompts that ask model to directly express the empathetic response based on the content and emotion without any reasoning steps, as shown in Fig~\ref{fig:model}.

\textbf{Zero-shot CoT}
Zero-shot CoT \cite{zhang2022automatic} uses a prompt to guide the LLM in generating reasoning steps for a task without prior task-specific examples. A traditional example of zero-shot CoT is the prompt ``Let's think step by step."
We further extend this approach by manually defining inferential steps (see Fig~\ref{fig:model}) and adding explanations for each step to enhance the model’s reasoning \cite{zhou2022least, wang2023plan}.

\textbf{Few-shot CoT}
Few-shot CoT \cite{wei2022chain} provides one or more task-specific examples that demonstrate step-by-step reasoning before solving a similar task. These examples are also be called rationale. We provide transcriptions and emotion labels for LLaMA2-7B-chat to generate a set of rationales using the zero-shot CoT prompt with manually defined steps. During inference, we randomly select one rationale from this set as the few-shot sample.

Since LLMs are highly sensitive to prompts, we employ the following prompt engineering techniques. We start with zero-shot prompts and progressively introduce additional steps, reasons, and few-shot examples to provide more context.
We evaluate each prompt on several samples of test set using our two-stage trained model. By iteratively refining the prompts, we identify the most effective prompt for each CoT type, leading to the results in Section~\ref{cot_result}.

\section{Experiments}

\subsection{Training Setups}
\label{setup-pt}
In Stage 1, we used 960 hours annotated speech of LibriSpeech dataset \cite{panayotov2015librispeech} for ASR tasks. In stage 2, we employ two labeled SER datasets: IEMOCAP \cite{busso2008iemocap} (excluding session 5), and MEAD \cite{wang2020mead}, along with 20\% of the LibriSpeech data replayed for ASR tasks. The multitask training ratio is 0.2, 0.3, 0.5 for ASR, SER and both.

The learning rate is 2e-4 and 5e-5 for stage 1 and 2, respectively. The number of trainable parameters is 79.7M (1.1\% of total parameters). Our training is trained on 2 NVIDIA A100 40G. Limited by the GPU memory, the batch size is set to 4 with a 8 gradient accumulation. We train stage 1 for 1 epoch on all 960 hours data and stage 2 for 10 epochs for combination of dataset mentioned before.
For inference, we apply a temperature of 0.7 and use top-p sampling with a probability threshold of 0.85.

\subsection{Test Setup}
To evaluate the generation capability of our model, we constructed a test set over IEMOCAP session 5. We select appropriate dialogue questions and balance the corresponding emotion labels to create a diverse and representative sample set. Additionally, to enrich the test cases, we generate an emotional speech test set with Alpaca 52k dataset \cite{alpaca} and Azure TTS API, referred to as AlpacaTTS. This dataset contains 6 emotion labels and questions including both objective factual inquiries and subjective generative prompts, providing a broad range of testing scenarios. We use GPT-4o to generate ground truth empathetic responses based on the question and emotion label. 
Furthermore, we manually verify all responses to ensure the quality of GT responses. 

We evaluate the ASR performance of our trained model using the Librispeech test-clean set and AlpacaTTS, measuring both word error rate (WER) and character error rate (CER). For SER evaluation, accuracy (ACC) is employed. The IEMOCAP session 5, which is excluded from the training process, and AlpacaTTS are used to assess SER performance.

\subsection{Baselines}

\textbf{Cascaded System} 
The speech dialog pipeline, with a ASR model like whisper \cite{radford2022whisper}, an emotion recognition model like emotion2vec \cite{ma2023emotion2vec}, and a text LLM to take transcribed text and emotion label to generate an response.

\textbf{SpeechLLM} 
We compare our approach with popular end-to-end speech LLMs such as SALMONN \cite{tang2023salmonn} and Qwen-Audio-Chat \cite{chu2023qwen}. Due to the unavailability of model weights and annotated dialogue datasets, we are unable to reproduce the results for E-CHAT and SpokenLLM mentioned in Section~\ref{introduction}, and thus these models are not included in our comparative analysis.

Since each model is optimized for different prompt styles, to ensure the generation performance, we use the official suggested prompt for each speechLLM. As for the cascade model, we use the same CoT prompt as our LPE.

\vspace{-1mm}
\subsection{Generation Evaluation Metrics}
To comprehensively assess the generation of our model, we employ both objective and subjective evaluation metrics on test set. 

The objective metrics include BLEU-1, BLEU-4 \cite{papineni2002bleu}, and BERTScore \cite{zhang2019bertscore}, which provide quantitative measures between generated response and ground truth response. BLEU-n scores are calculated to evaluate the n-gram overlap between the generated responses and reference answer. BERTScore is utilized to assess the semantic similarity between the generated responses and the reference text, leveraging contextual embeddings from BERT \cite{devlin2018bert}.

Following main stream evaluation metric for natural language generation (NLG) \cite{liu2023g, fu2023gptscore, gao2024llm}, we provide question and emotion labels in text form as conditions, with generated text response as input to GPT-4o \cite{achiam2023gpt, openai2024gpt4o} to score the generated responses from 1-5 based on empathetic quality, content helpful and generate clarity.
This subjective evaluation offers insights into the model's ability in three aspect: 1) Content. How well does the response stay on-topic and provide an appropriate reply to the question. 2) Empathy. How effectively does the response acknowledge and respond to the emotional state of the questioner. 3) Clarity. How grammatically correct and state clear is the response. Moreover, we also use GPT-4 to perform a win-rate comparison to explore the general performance of empathetic response between two models.

\begin{table}[t]\scriptsize
  \caption{ASR and SER results on different datasets}
  \label{tab:obj1}
  \centering
  \begin{threeparttable}
\begin{tabular}{l|cc|c|c|c}
\toprule
\multirow{2}{*}{}  & \multicolumn{2}{c|}{\textbf{LibriSpeech}} & \multicolumn{2}{c|}{\textbf{AlpacaTTS}} & {\textbf{IEMOCAP}}  \\
                 & \textbf{WER} $\downarrow$ & \textbf{CER} $\downarrow$ &  \textbf{WER} $\downarrow$ & \textbf{ACC} $\uparrow$ & \textbf{ACC} $\uparrow$ \\
\midrule
LPE w/ stage1      & 1.357 & 1.491  & 0.611 & 42.79 & 45.31\\
\quad + w/ stage2        & 1.359 & 1.594  & 0.600 & 70.39 & 73.64\\
\bottomrule
\end{tabular}
\end{threeparttable}
\vspace{-2mm}
\end{table}

\begin{table}\footnotesize
  \caption{Objective results on generated response}
  \label{tab:obj2}
  \centering
  \begin{tabular}{lcccccc}
    \toprule
                 & \textbf{BLEU$_1$} $\uparrow$ & \textbf{BLEU$_4$} $\uparrow$ & \textbf{ BERTScore$_{f1}$} $\uparrow$\\
    \midrule
    LPE             & \textbf{16.13} & \textbf{2.85} & \textbf{82.9}\\
    \quad -w/o stage 2      & 8.43 & 1.12 & 81.1\\
    SALMONN         & 3.54 & 1.03 & 80.6\\
    Qwen-Audio-Chat & 8.29 & 0.87 & 80.0\\
    \bottomrule
  \end{tabular} 
  \vspace{-3mm}
\end{table}

\section{Results}
\subsection{Two-stage Training Analysis}
\label{objective}
We first apply experiments on ASR and SER to evaluate the effectiveness of two stage training. We use prompt to ask LLM generate the transcription and emotion labels for input speech.
As shown in Table ~\ref{tab:obj1}, our method effectively trains LLMs to understand both textual content and emotional labels, achieving a competitive ASR WER and SER ACC across datasets. Specifically, stage 2 contributes to a 30\% improvement in emotion classification without compromising WER, proving the effectiveness of mutlitask training.

\begin{figure}[h]
  \centering
  \includegraphics[width=0.9\linewidth]{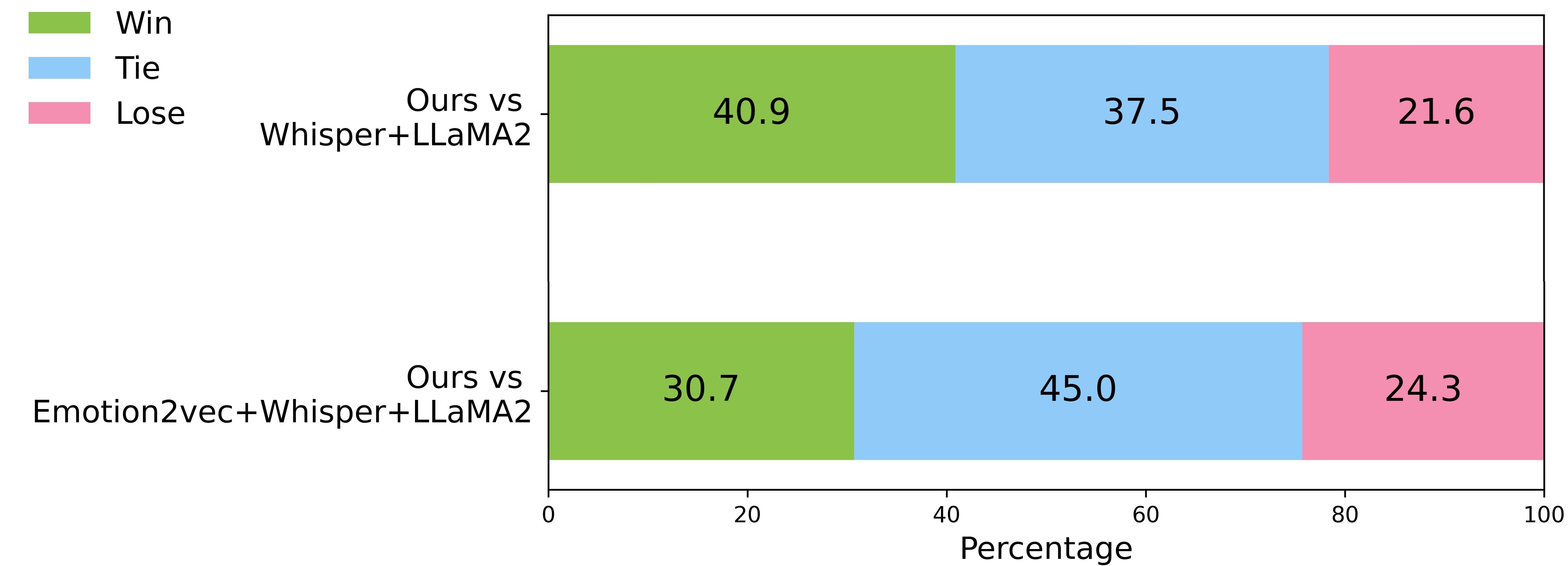}
  \caption{Win rate compared with cascade model.}
  \label{fig:winrate}
  \vspace{-1mm}
\end{figure}

\subsection{Generation Analysis}
\label{alignment}
Table~\ref{tab:obj2} shows the objective metrics comparing each model's response to the ground truth. Our LPE outperforms all other speech-based LLMs in all three scores. As noted earlier, traditional evaluation metrics are limited in assessing generation quality, so we conduct subjective evaluation following the popular mainstream methods.
The results in Table~\ref{tab:sub} further analyze the response quality in terms of content, empathy, and clarity. We report the average score for each dimension, along with the standard deviation ($\pm$). Compared to SALMONN and Qwen-Audio-Chat, LPE demonstrates a well-rounded performance across all aspects, with clear advantages in empathy and content richness, highlighting its effectiveness in providing both empathetic and helpful responses.

In Table~\ref{tab:sub}, we evaluate cascaded systems combining Emotion2vec, Whisper, and LLaMA2. Our LPE performs on par with the SOTA cascaded model in content and clarity scores, while slightly surpassing it in empathy, with minimal score differences (within 0.1).  To better assess their strengths and weaknesses, we use the win-rate metric, a widely accepted evaluation method in NLG.
As illustrated in Fig~\ref{fig:winrate}, LPE achieves higher win rates, demonstrating superior emotional understanding and response capabilities. Notably, LPE outperforms the Emotion2vec + Whisper + LLM pipeline, a leading SOTA approach. This may be due to: 1) Emotion labels: LPE provides richer emotional information compared to Emotion2vec’s single-label output (Fig~\ref{fig:model}), resulting in more nuanced responses. 2) Prompt engineering: Optimizing prompts for LLMs could further improve cascaded model performance.

\begin{table}\footnotesize
  \caption{Subjective results on generated response}
  \label{tab:sub}
  \centering
  \begin{tabular}{lccc}
    \toprule
         & \textbf{Content} & \textbf{Empathy} & \textbf{Clarity}\\
    \midrule
    LPE             & 3.99 $\pm$ 0.54 & \textbf{3.72 $\pm$ 0.90} & 4.97 $\pm$ 0.17\\
    SALMONN         & $2.12 \pm 0.32$ & $1.34 \pm 0.56$ & $4.94 \pm 0.23$\\
    Qwen-Audio-Chat & $2.00 \pm 0.46$ & $1.63 \pm 0.96$ & \textbf{4.98 $\pm$ 0.14}\\
    \midrule
    Whisper+LLaMA2  & 3.97 $\pm$ 0.46 & $2.87 \pm 0.96$ & $4.95 \pm 0.20$\\
    \quad + w/ emotion2vec & \textbf{4.01 $\pm$ 0.34} & 3.64 $\pm$ 0.52 & \textbf{4.98 $\pm$ 0.14}\\
    \bottomrule
  \end{tabular}
  \vspace{-1mm}
\end{table}

\begin{table}
\vspace{-1mm}
    \caption{Subjective evaluation on different CoT prompts}
    \label{tab:cot}
    \centering
    \begin{tabular}{lcc}
    \toprule
        \textbf{Prompt} & \textbf{Content} & \textbf{Empathy} \\
    \midrule
        No CoT                            & 2.14 & 1.35 \\
        Zero Shot CoT                     & 3.37 & 3.14 \\
        \quad + w/ steps \textbf{(Ours)}  & \textbf{3.99} & \textbf{3.72} \\
        \quad + w/ reasoning              & 2.81 & 2.97 \\ 
        Few shot CoT                      & 1.79 & 3.43 \\
    \bottomrule
    \end{tabular}
    \vspace{-1mm}
\end{table}

\subsection{CoT Analysis}
\label{cot_result}
We evaluate the CoT prompts mentioned in Section~\ref{cot} on our trained model. The content and empathy scores are shown in Table~\ref{tab:cot}. The results show that proper CoT prompt can guide model to generate empathetic response even never SFT with QA dataset. Among the different CoT types, the zero-shot CoT with predefined steps yields the best performance. Meanwhile adding additional reasoning to these steps leads to a decrease in both content and empathy scores. Surprisingly, the commonly used few-shot CoT performs bad on our model with the lowest content score.

We observe several key findings base on the further case study.
1) Impact of Prompt Length: The length of the CoT prompt appears to significantly influence generation performance. For instance, longer prompts, such as reasoning zero-shot CoT and few-shot CoT, tend to increase the rate of containing repetitive or meaningless words causing a low content score. On the other hand, a concise CoT that provides necessary steps performing more effectively. 2) Importance of Steps: From the result of original zero-shot CoT, we notice that the text LLM sometimes struggles to infer proper steps on a speech dialogue tasks, causing the missing of key steps during generation.
In contrast, the response quality increases a lot when we add a manually defined inferential steps.
3) Challenges with Few-shot CoT: Unlike its strong performance in text-based LLMs, few-shot CoT prompt yields a low content score, as it tends to generate responses that overly focus on textual content rather than input speech. Thus, although the empathy score is relatively high, it is not related to actual input speech causing a low content score. Based on these findings, we finally design a zero-shot CoT characterized by clearly defined steps as our prompt.

\begin{table}
    \caption{Ablation study with subjective evaluation}
    \label{tab:ablation}
    \centering
    \begin{tabular}{lcc}
    \toprule
        \textbf{Method} & \textbf{Content} & \textbf{Empathy} \\
    \midrule
        LPE              & 3.99 & 3.72 \\
        - w/o stage2            & 1.97 & 1.35\\
        - w/o multitask         & 3.21 & 2.50 \\
        - w/o emotion loss      & 4.00 & 3.31 \\
    \bottomrule
    \end{tabular}
    \vspace{-2mm}
\end{table}

\subsection{Ablation Study}
\label{ablation}
In this section, we conduct an ablation study to evaluate the contributions of our model's components. First, we remove Stage 2 and observe a significant drop in both objective and subjective metrics (Tables ~\ref{tab:obj2} and ~\ref{tab:ablation}), highlighting its critical role in refining empathetic responses. We observe that, without Stage 2, the model often generates transcriptions instead of responses, likely due to overfitting on the ASR task during Stage 1. Stage 2 mitigates this issue, improving performance.
Next, we assess multitask training by freezing the subsampler and training the multi-scale adapter with single SER task. While this slightly reduces overfitting, the model still produces many transcribed responses, leading to a significant performance drop. This underscores multitask training's role in reducing overfitting and helping the model learn emotional features.
Finally, we evaluate emotion loss and find it enhances the model's ability to capture emotional nuances, further boosting performance.

\section{Conclusions}
In this work, we propose an novel framework, referred as LPE, which eliminates the reliance on emotional spoken QA datasets, effectively addressing the major bottleneck in the development of empathetic spoken dialogue systems. Our approach utilizes a two stage training process to train text-based LLM to listen speech and perceive the emotion. Then, we apply CoT with manually defined inference steps to guide model listen the content, perceive the emotion and express the empathetic response without SFT on emotional QA data. Experimental results demonstrate that our LPE framework can produce high quality empathetic response, proving the effectiveness of our proposed approach.

\bibliographystyle{IEEEtran}
\bibliography{IEEEabrv, refs}

\end{document}